\def\year{2018}\relax
\documentclass[twocolumn]{article}
\usepackage{multicol,lipsum}
\usepackage{aaai18}
\usepackage{times}
\usepackage{helvet}
\usepackage{courier}
\usepackage{graphicx}
\usepackage{subfigure}
\usepackage{amssymb}
\usepackage{amsmath}
\usepackage{booktabs}
\usepackage{bm}
\usepackage{mathtools}
\usepackage{threeparttable}
\usepackage[american]{babel}
\usepackage{color}
\usepackage{threeparttable}
\usepackage{appendix}
\usepackage{comment}

\usepackage[left]{lineno}
\usepackage{blindtext}
\usepackage{amsmath}

\newcommand{\bs}{\boldsymbol}

\begin{document}
%
%
\title{Rocket Launching: A Universal and Efficient Framework for Training Well-performing Light Net}
\author{Guorui Zhou,\textsuperscript{1}\thanks{Correspondence author is Guorui Zhou. The source code is available at https://github.com/zhougr1993/Rocket-Launching.} Ying Fan,\textsuperscript{1} Runpeng Cui,\textsuperscript{2} Weijie Bian\textsuperscript{1}\\
    ¦   {\bf \Large Xiaoqiang Zhu\textsuperscript{1} \and Kun Gai\textsuperscript{1}}\\
    ¦   \textsuperscript{1}Alibaba Inc, Beijing, China\\
        \textsuperscript{2}Department of automation, Tsinghua university\\
\{guorui.xgr, fanying.fy, weijie.bwj, xiaoqiang.zxq\}@alibaba-inc.com, cuirunpeng@gmail.com, jingshi.gk@taobao.com\\
}
\maketitle


\begin{abstract}
 Models applied on real time response tasks, like click-through rate (CTR) prediction model,
 require high accuracy and rigorous response time.
 Therefore, top-performing deep models of high depth and complexity are not well suited for
 these applications with the limitations on the inference time.
 In order to get neural networks of better performance given the time limitations,
 we propose a universal framework that exploits a booster net to help train the lightweight net for prediction.
 We dub the whole process rocket launching, where the booster net is used to
 guide the learning of our light net throughout the whole training process.
 We analyze different loss functions aiming at pushing the light net to behave similarly to the booster net.
 Besides, we use one technique called gradient block to improve the performance of light net and booster net further. Experiments on benchmark datasets and real-life industrial advertisement data
 show the effectiveness of our proposed method.
\end{abstract}
\section{Introduction}

Deep networks have achieved state-of-the-art results in many areas, such as computer vision~\cite{huang2016densely} and nature language processing~\cite{bahdanau2014neural}.
From AlexNet~\cite{krizhevsky2012imagenet} to recently proposed DenseNet~\cite{huang2016densely},
better performances are accompanied with deeper and wider networks and more complex and adaptable structures.
A more complex structure of neural networks means longer inference time,
which is not tolerated in industrial environment.
Networks mentioned above only consider the evaluation criterion of accuracy,
while neglect the necessity of real-time response in industrial applications.

At the same time, some nets like DIN~\cite{zhou2017deep} and wide~\&~deep model~\cite{cheng2016wide} get more and more attention. These nets share some characteristics: nets are shallow,
layers are very simple and with less computation cost.
In industrial applications, e.g.~online advertising systems, models have to make prediction of hundreds of advertisements for one user in several milliseconds, which restricts the complexity of model.
Only simple and shallow structure meets the stringent response time requirements in industry.

Accuracy and latency are the two points that we pay attention to.
In general, there are two solutions to reduce runtime complexities while keeping a decent performance.
Some works make use of factorizing or compressing to directly simplify the computation,
such as matrix SVD~\cite{denton2014exploiting},
MobileNet~\cite{howard2017mobilenets}, and ShuffleNet~\cite{zhang2017shufflenet}.
Other approaches adopt the teacher-student strategy. They use light networks with fewer layers and parameters to decrease the inference time, while the light nets are trained helped by a complicated teacher network that trained in advance,
like knowledge distillation~\cite{hinton2015distill} and FitNet~\cite{Romero2014fitnet}.
These teacher-student methods decrease the runtime complexities,
and can be further combined with approaches of the first category.
In this work, we propose a novel universal framework to train decent small networks,
motivated by the potential of teacher-student methods.

In this work, we develop a novel network training process dubbed rocket launching.
The light net is the target network for inference, the booster relates to the deeper and more complex network from the architecture.
Both the light and the booster net compose the architecture of rocket network.
At the training stage, the light and booster networks are trained simultaneously on the same task.
Besides, the light net also keeps getting knowledge learned by the booster through the optimization of the hint loss,
which is included in the objective function to make both nets have similar behaviour during training.
The booster guides the optimization of the target light network along all the training process.
At the inference stage, only the trained light network is used.
Different from previous teacher-student methods~\cite{hinton2015distill,Romero2014fitnet},
we make the light model share some lower layers with the cumbersome one
and train them simultaneously.

In this paper, we propose a universal method aiming to obtain a well-behaved light net considering limitations on inference time.
Our method is suitable to many different network structures.
In brief, our contributions can be summarized as follows:
\begin{itemize}
\item We propose a novel universal training process called rocket launching, which makes use of the booster net to supervise the learning of the light network through whole training process.
    We show that a light model can be trained to perform close to deeper and more complex models in experiments.
\item We analyze different hint loss functions to transfer the information from booster to the light net.
\item In order to push light net to be close to booster net, we use gradient block technique to cancel the effect of hint loss's back-propagation on layers of the booster, which gives booster net more freedom to update its parameters based on ground truth and improve the performance further.
\end{itemize}

Our method achieves the state-of-the-art results on publicly available benchmarks
as well as industrial dataset.
It is notable that our method performs better than other teacher-student approaches.
Experimental results present that the performance can be further improved when combining other teacher-student approaches with our framework.

The remainder of this paper starts from a summary of related work.
Then we introduce our approach, followed by experiments and conclusions.
\section{Related work}
Deep neural networks have drawn increasing attention in recent years due to their overwhelming performance
on many research areas.
One main trend of network structure design is to develop neural networks with
larger depth, more parameters and higher complexity to achieve better performance~\cite{Simonyan2015verydeep,szegedy2015going,he2016deep,Zagoruyko2016wide}.
However, these top-performing networks with high complexity will result in time consuming systems
at the inference phase.
Therefore, they are not well suitable for applications with inference time limitations.

There have been some explorations of model compression by directly simplifying the computation or pruning of the original neural operations. Denton et al.~\cite{denton2014exploiting} use SVD to approximate the convolutional operations in deep CNNs. MobileNets~\cite{howard2017mobilenets} are based on a streamlined architecture that uses depthwise separable convolutions to build lightweight deep neural networks. ShuffleNet~\cite{zhang2017shufflenet} uses pointwise group convolution and channel shuffle to reduce computation cost. ThiNet~\cite{luo2017thinet} uses statistic information from next layer to prune filters which accelerates CNN models while maintaining accuracy.

Besides designing delicate net structure, light net can get more information from extra pre-trained model during training phrase. This idea has been emphasized in Learnware~\cite{zhou2016learnware}. There have been some attempts adopting a teacher-student strategy, where a more complex teacher network is employed to teach a lightweight student network on a given task.
The teacher network helps the student net to get a decent performance at the inference phase.
Bucilu\v{a} et al.~\cite{Bucilua06compression}
improve compression model, which pioneers this type of learning process. They expound that the knowledge of a large ensemble of models could transfer to a single small model, they use a large ensemble of models to label large amounts of unlabel data, then use the data labeled by the ensemble models to train small model. Furthermore,
Ba et al.~\cite{ba2014deep}
train a wider and shallower net called student net to mimic the big model called teacher net via regressing logits before the softmax layer with $\ell_2$ loss. They think that matching logits could get more information than the hard label that provided by the cumbersome model.
Hinton et al.~\cite{hinton2015distill}
point out that identifying knowledge in a trained model with the learned parameter value is hard. Instead, they make use of one abstract view of the knowledge that is the learned mapping from input vectors to output vectors, they propose the strategy of knowledge distillation which uses the class probabilities produced by the cumbersome model as "soft targets" for training the small model. They prove that they are the general version of matching logits which uesd by~\citeauthor{ba2014deep}\shortcite{ba2014deep}.

Besides using the output of the teacher network, people try to get more supervised information from the teacher. FitNets~\cite{Romero2014fitnet} use not only the outputs but also the intermediate representations learned by the cumbersome model as the hint to supervise the training process. Zagoruyko et al.~\cite{Zagoruyko2016attenion} use attention as a mechanism of transferring knowledge from one network to another. By properly defining attention for convolutional neural networks, they improve the performance of a student CNN by forcing it to mimic the attention maps of a powerful teacher network.

In previous teacher-student approaches, the cumbersome teacher networks are trained in advance.
Instead of only transferring the final stationary outputs of the pretrained model,
we let the booster model guide the whole training process of light net in rocket launching.
We think that the knowledge learned by the cumbersome model exists not only in the final outputs,
but also in the full learning process.
The light model gets not only the difference between the target and the temporary outputs, but also the possible path towards the final target provided by a complex model of more learning capability.
Another difference of our approach is that the part parameters of the light model and the booster are shared in our framework. We adopt the parameter sharing scheme since the lower-level representation of the same task should be universal. In the proposed architecture, the booster has a much deeper specific layers to ensure the capability to guide the light model to learn the task better.

Training several nets together is often applied on multi input scenes~\cite{Andrew13dcca,Bromley1993siamese} or semi-supervised task~\cite{laine2016temporal}. Parameters sharing has also been used in multi-task~\cite{He2017maskrcnn}. However, to the best of our knowledge, there is no attempt on using these techniques to train small net to get better performance. We are the first to utilize these schemes in model compression attempts, and results in experiments present the effectiveness of our method.

\section{Our approach}\label{sec:opa}

In this section, we will describe our proposed rocket net training process in detail.
We will further analyze the highlights of our method and compare different hint loss functions.

\subsection{The sketch of our method}

\begin{figure}[!t]
\centering
\includegraphics[height=80mm ]{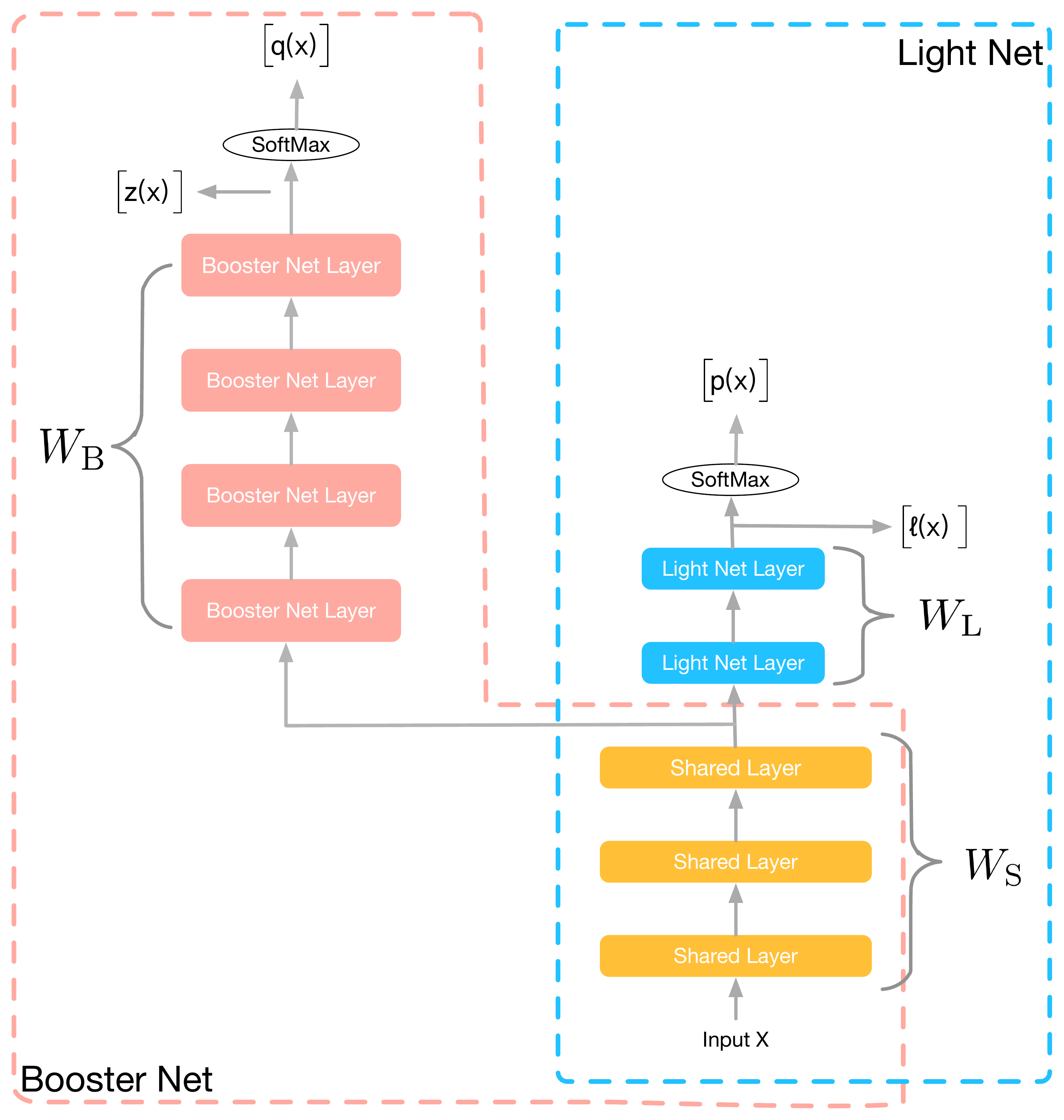}
\caption{{Whole Net Structure, blue dashed circle represents light net, pink dashed circle represents booster net. Yellow layers are  shared by light net and booster net.}}
\label{fig:gm}
\end{figure}

Fig.~\ref{fig:gm} represents the general structure of our architecture,
it is composed of two parts: the light net and the booster net. These two networks share some lower-level layers (annotated in yellow), and they both have their specific layers for the learning and prediction on the same task.

%

We let $\bs{x}$ and $\bs{y}$ denote the inputs and one-hot ground truth labels of our neural architecture.
Let $\mathrm{L}$ be the light net with an output softmax as
$\bs{p}(\bs{x})=\mathrm{softmax}(\bs{l}(\bs{x}))$,
where $\bs{l}(\bs{x})$ is the weighted sum before the softmax activation.
Parameters for the light net consist of two components: parameters in shared layers $\bs{\mathrm{W}}_\mathrm{S}$ and
parameters in its lightweight particular layers for prediction $\bs{\mathrm{W}}_\mathrm{L}$.
We let $\mathrm{B}$ denote the booster network with shared parameters $\bs{\mathrm{W}}_\mathrm{S}$ and
its particular weights $\bs{\mathrm{W}}_\mathrm{B}$ to get the final output.
Similar to the light net, we have $\bs{q}(\bs{x})=\mathrm{softmax}(\bs{z}(\bs{x}))$
as the output softmax for the booster,
where $\bs{z}(\bs{x})$ is the weighted sum before the softmax activation.
We expect that the light net is trained similar to the true labels $\bs{y}$,
as well as approximate to the knowledge learned by the booster net with much more representation capability.
To solve this problem, we take hint loss in the training objective in order to
convey knowledge from the booster net to the light net.
The objective function for rocket launching is defined as follows:
\begin{eqnarray}
\begin{split}
\mathcal{L}(\bs{x};\boldsymbol{\mathrm{W}}_\mathrm{S},\boldsymbol{\mathrm{W}}_\mathrm{L},\boldsymbol{\mathrm{W}}_\mathrm{B}) = \ &\mathcal{H}(\bs{y}, \bs{p}(\bs{x}))+\mathcal{H}(\bs{y}, \bs{q}(\bs{x}))\\
&+\lambda\|\bs{l}(\bs{x}) - \bs{z}(\bs{x})\|_2^2,
\end{split}
\label{eq:obj}
\end{eqnarray}
where the last term is the hint loss function as the mean square error (MSE)
between the logits $\bs{z}(\bs{x})$ and $\bs{l}(\bs{x})$,
$\mathcal{H}(\bs{p}, \bs{q})=-\sum_i p_i\log q_i$ is the cross-entropy,
$\lambda$ is the parameter to balance the cross-entropy and the hint loss.
Here we use the cross-entropy terms for the booster and light nets to learn the true labels,
and use hint loss function to exploit the knowledge learned by the booster to guide the learning process
of the light network.

\subsection{Characters of our method}\label{subsec:character}
There are some highlights in our method, which have notable effects on the training process
and distinguish our method from other teacher-student approaches.
\subsubsection{Parameter sharing}
In our approach, the light net shares parameter with the booster net.
This scheme helps the light net get direct thrust from the booster,
which pushes it get better performance.

The technique of parameter sharing is not new in deep learning.
In the area of computer vision, it is a common scheme to train deep convolutional neural networks in a multi-task manner.
We assume that these tasks can be built on some shared low-level representations of the images.
Given this assumption, we could reduce the parameters in neural network and improve its generalization capability.
It is noticeable that, in industrial applications, e.g.~CTR prediction,
reusing the embedding layers from other tasks helps a new task converge more easily and get a better performance.

\subsubsection{Simultaneous training}
 In most teacher-student methods, the teacher network is trained on the target database in advance,
 and its parameters are fixed when guiding the training process of the student net.
 Different from these approaches, we have the light and booster nets trained simultaneously,
 the whole learning process of the target light net is guided by the booster net.
 The light model can learn from not only the difference between the target and its temporary outputs, but also the possible path towards the final target provided by a complex model with more learning capability.

 Notice that instead of training the teacher and student nets separately,
 the whole training time of our proposed architecture is shortened.
 Therefore, the compressed model can be trained more efficiently to meet the requirement in industrial applications
 that the inference model be updated frequently.

\subsubsection{Hint loss functions}
In our approach, we transfer the booster net's knowledge to the light net by minimizing the hint loss.
Several different hint loss functions are considered in this work:
\begin{itemize}
\item MSE of final softmax: $\mathcal{L}_\mathrm{MSE}(\bs{x})=\|\bs{p}(\bs{x}) - \bs{q}(\bs{x})\|_2^2$,
\item MSE of logits before softmax activation, which is also adopted in SNN-MIMIC~\cite{ba2014deep}:
    $\mathcal{L}_\mathrm{mimic}(\bs{x})=\|\bs{l}(\bs{x}) - \bs{z}(\bs{x})\|_2^2$,
\item knowledge distillation~\cite{hinton2015distill}: $\mathcal{L}_\mathrm{KD}(\bs{x})=\mathcal{H}(\frac{\bs{p}(\bs{x})}{T}, \frac{\bs{q}(\bs{x})}{T})$,
    where $T$ is the temperature.
\end{itemize}

For the MSE of final softmax $\mathcal{L}_\mathrm{MSE}$,
we have the derivative of the hint loss with respect to $\bs{l}_i(\bs{x})$:
\begin{eqnarray}
\begin{split}
\frac{\partial\mathcal{L}_\mathrm{MSE}(\bs{x})}{\partial \bs{l}_i(\bs{x})} &= 2\bs{p}_i(\bs{x})\Big[\bs{p}_i(\bs{x}) - \bs{q}_i(\bs{x}) \\
&+ \sum\nolimits_k \bs{p}_k(\bs{x})(\bs{q}_k(\bs{x}) - \bs{p}_k(\bs{x}))\Big].
\end{split}
\end{eqnarray}
Notice that the gradient is proportional to the prediction outputs of the light net.
If $\bs{l}_i(\bs{x})$ is very negative, causing $\bs{p}_i(\bs{x})$ close to zero and the gradient to vanish,
the MSE of final softmax may fail to learn the difference in outputs,
even when the light net makes radically different outputs from the booster net.

SNN-MIMIC learning~\cite{ba2014deep} uses the formulation of $\mathcal{L}_\mathrm{mimic}$
between the teacher and student networks.
We have the derivative w.r.t. $\bs{l}_i(\bs{x})$:
\begin{eqnarray}
\frac{\partial\mathcal{L}_\mathrm{mimic}(\bs{x})}{\partial \bs{l}_i(\bs{x})} = \bs{l}_i(\bs{x}) - \bs{z}_i(\bs{x}).
\end{eqnarray}
We observe that the update directly reduces the difference between the logits before softmax,
which prevents the problem of gradient vanishing with $\mathcal{L}_\mathrm{MSE}$.
Experimental results also present that training with $\mathcal{L}_\mathrm{mimic}$
achieves the best performance among these different hint loss formulations.

Knowledge distillation~\cite{hinton2015distill} uses cross-entropy to restrict the probability outputs of two models.
In their work, a temperature $T$ is introduced to produce a softer probability distribution among classes.
They think that knowledge distillation is the general case of matching logits.
They prove that with a high temperature, the gradient w.r.t.~$\bs{l}_i(\bs{x})$ is:
\begin{eqnarray}
\frac{\partial\mathcal{L}_\mathrm{KD}(\bs{x})}{\partial \bs{l}_i(\bs{x})}\approx\frac{1}{NT^2}(\bs{l}_i(\bs{x}) - \bs{z}_i(\bs{x})),
\end{eqnarray}
where $N$ is the number of classes, and approximation $\mathrm{e}^{\bs{l}_i(\bs{x})/T}\approx 1 + \bs{l}_i(\bs{x})/T$ is used.
Their approximation neglects the term $(\bs{l}_i(\bs{x})/T)^2$ in Taylor series
when the temperature is high enough compared with the magnitude of the logits.
Notice that the approximate gradient $\frac{1}{NT^2}(\bs{l}_i(\bs{x})-\bs{z}_i(\bs{x}))$ is the same order of infinitesimal to the neglected term $(\bs{l}_i(\bs{x})/T)^2$, this approximation may also cause a negligible gradient.
But we approve the temperature's effect that it can soften class probability,
which makes the distillation pays more attention to matching the negative logits below the average.
In practice, Hinton et al.~\cite{hinton2015distill} suggest that intermediate temperatures work best,
which ignores the very negative logits that might be noisy.
While in this work, we find that the optimization of all the logits' difference in our framework outperforms using the formulation of $\mathcal{L}_\mathrm{KD}$.
We think that some very negative logits may convey useful knowledge acquired by the cumbersome net,
which helps the student network to get a better performance.

\subsubsection{Gradient block}
\begin{figure}[!t]
\centering
\includegraphics[width=70mm ]{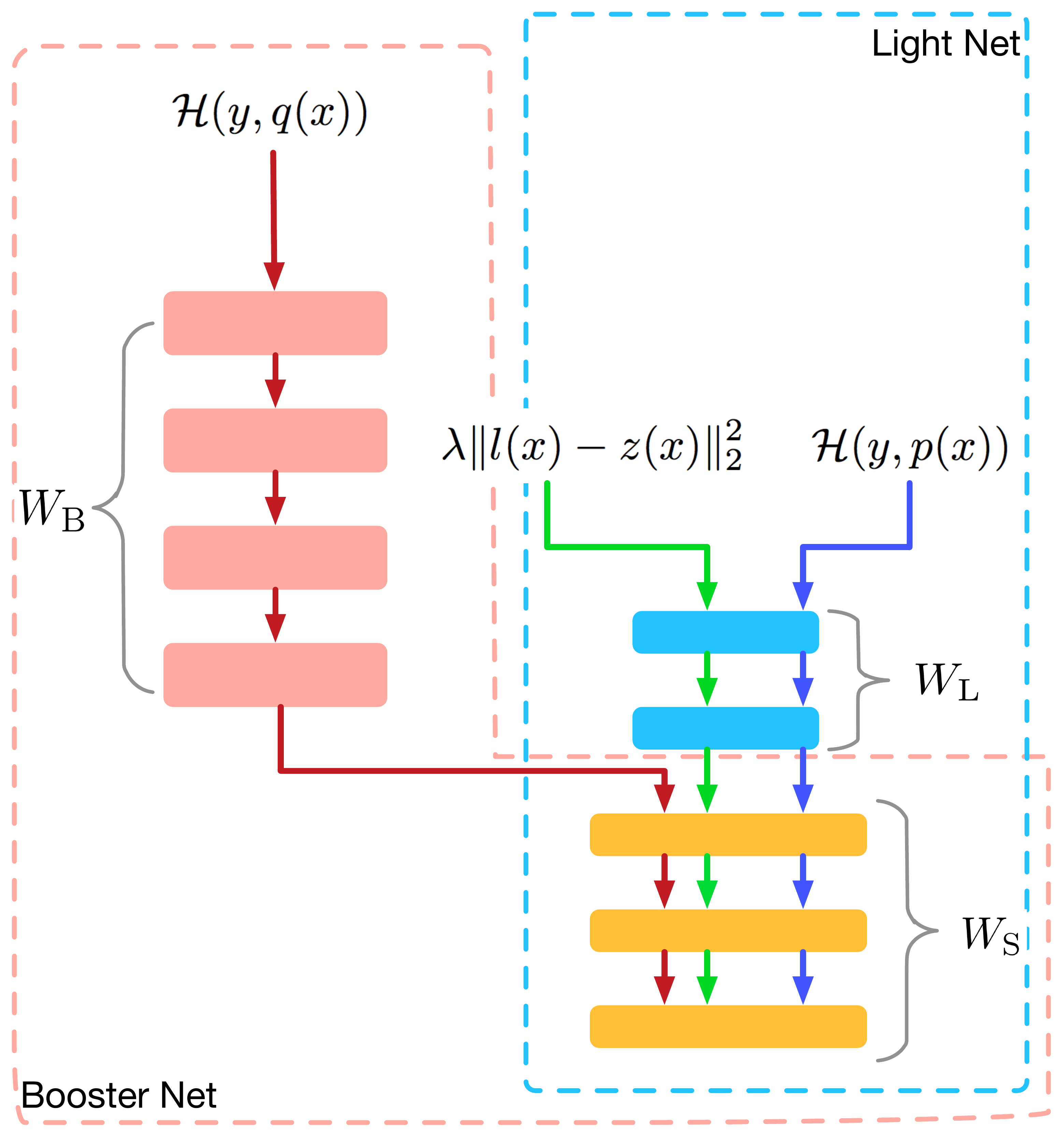}
\caption{{Gradient backward with the gradient block scheme.}}
\label{fig:gb}
\end{figure}
In our proposed training process, the light net shares parameters and is trained together
with the booster net.
This simultaneous training scheme has an inevitable effect on the performance of the booster network.
Using both cross-entropy $\mathcal{H}(\bs{y},\bs{q}(\bs{x}))$ and hint loss as the objective to update booster's parameters will make the categorical outputs of the booster strongly affected by those of the light net, and hinder the booster from learning on the task directly.
Since the learning capability of the light model is limited,
the performance of the booster net will be inevitably deteriorated.
Notice that the light model learns from the knowledge conveyed by the booster net during training,
this deterioration in the booster model's learning will further
diminish the learning potential of the light network.

In order to solve this problem, during the training process, we develop the gradient block scheme to prevent the booster model
from minimizing the hint loss objective. As we can see from Fig.~\ref{fig:gb}, during the back-propagation of the hint loss term, we fix the gradient of booster net's specific parameters~($\bs{\mathrm{W}}_\mathrm{B}$), and use this moment booster net's probability as target to supervise light net's study.

This operation makes the specific parameters $\bs{\mathrm{W}}_\mathrm{B}$ in booster net away from the effect given by
the light model, thus the booster can directly learn from the ground truth labels to achieve its best performance.
For the light net, the parameters are normally updated to optimize the objective function in Eq.~\ref{eq:obj}.
Both the supervisory information and the booster's knowledge are the targets for the light model to learn from.

\begin{table*}[!t]
  \caption{Comparisons of classification performance(test error) on CIFAR-10}
  \label{tab:cifar10}
\centering
{
\centerline {
\begin{threeparttable}
\begin{tabular}{ccccccccc}
\toprule
light & booster & base\tnote{1} & AT & KD & rocket\tnote{2} & rocket+KD\tnote{3} & booster\tnote{4} & booster only\tnote{5}\\
\midrule
WRN-16-1,~0.2M(b) & WRN-40-1,~0.6M & 8.77 & 8.25 & 8.39 & 7.87 & 7.52 & 6.64 & \textbf{6.58} \\
WRN-16-2,~0.7M(b) & WRN-40-2,~2.2M & 6.31 & 5.85 & 6.08 & 5.67 & 5.64 & 5.20 & \textbf{5.23} \\
WRN-16-1,~0.2M(a) & WRN-40-1,~0.6M & 8.69 &  -\tnote{6}   & 8.34 & 7.85 & 7.51 & 7.27 & \textbf{6.58} \\
\bottomrule
\end{tabular}
\begin{tablenotes}[para,flushleft]
        \item[1] base means WRN-16 trains individually.
        \item[2] rocket means light net's result in rocket launching.
        \item[3] rocket+KD means light net's result using rocket launching combined with KD.
        \item[4] booster means booster net's result in rocket launching.
        \item[5] booster only means WRN-40 trains individually.
        \item[6] WRN-16-1,~0.2M(b) can't be applied on AT directly, so we did not report this result.
      \end{tablenotes}
    \end{threeparttable}}}
\end{table*}

\section{Experiments}
In this section, we evaluate our rocket launching on several classification datasets and a real advertisement database from a Chinese leading e-commerce site. Experimental results present that our proposed approach achieves notable
improvements in the light net's performance and outperforms other teacher-student methods.
In experiments on public benchmarks, we compare our method with knowledge distillation~(KD)~\cite{hinton2015distill} and attention transfer~(AT)~\cite{Zagoruyko2016attenion}.

\subsection{Experiments on CIFAR-10}
The CIFAR-10 dataset~\cite{krizhevsky2009learning} consists of $32\times32$ color images
from $10$ class.
These images are split into $50,000$ training samples and $10,000$ testing samples.
We preprocess the data with the same operations as in~\citeauthor{Zagoruyko2016attenion}\shortcite{Zagoruyko2016attenion}.
All the experiments are repeated $3$ times with different seed,
and we take the median of error rates as the final results. All the experiments, we use the same learning rate tuning and epochs as in~\citeauthor{Zagoruyko2016attenion}\shortcite{Zagoruyko2016attenion}. We set the initial learning rate to be $0.1$ with momentum to be $0.9$, while we drop learning rate by $0.2$ at $[60,120,160]$ epochs and train for total $200$ epochs.

We employ wide residual net~\cite{Zagoruyko2016wide} to be the instantiation of rocket launching
on CIFAR-10 datasets. 
Wide residual net (WRN) has three groups of block, each block has two convolutional layers with larger width
in contrast with the original ResNet.
The wider layers are accompanied with more parameters, which could offer more representation capability.
Fig.~\ref{fig:structure}(a) shows the schematics of the rocket net structure based on wide residual networks.
Layers in red are shared by the light net and the booster. As we can see, sharing layers~(layers in red) are in the lower group of wide residual net. The yellow part is the specific structure designed for light net to make prediction.
The blue part is the specific layers of the booster, which is removed at inference phrase.
Attention transfer~(AT) uses teacher net's output activations of each group of residual blocks to supervise student net's each group's activations. In order to compare with AT fairly, we design another sharing way. As Fig.~\ref{fig:structure}(b) shows, the light net shares some lower blocks with the booster in each group.

%
%


We explore rocket launching on light and booster net with different network depths and widths~(e.g.~WRN-16-1(a),0.2M means wide residual network with depth of $16$ and widening factor of $1$, using the layer sharing way like Fig.~\ref{fig:structure}(a), its parameters' size is $0.2$M).
As shown in Table~\ref{tab:cifar10}, our approach achieves consistently notable improvement compared to the base light net with different experimental settings.
Taking the first line of Table~\ref{tab:cifar10} as example, using the same WRN-16-1(b) net structure, our rocket launching get $0.9\%$ improvement compared with this net trained individual.
We also observe that our approach outperforms other teacher-student methods,
such as knowledge distillation~(KD)~\cite{hinton2015distill} and attention transfer~\cite{Zagoruyko2016attenion}. It's notable that benefitting from
the structure characteristic of residual net, the way of sharing shown in Fig.~\ref{fig:structure}(b) still obtains decent result.

Besides comparing with other approaches, we also try to combine KD with our method by adding $\mathcal{L}_\mathrm{KD}$ to the objective in Eq.~\ref{eq:obj}. It's notable that we use probability that pre-trained by booster net in $\mathcal{L}_\mathrm{KD}$, which means light net can also obtain additional guidance from a pre-trained booster network. From Table~\ref{tab:cifar10}, we see that the performance can be further improved with the application of KD,
which means our rocket launching has different effect on the light net with KD.
The light net benefits from both the supervisory information brought by the pre-trained teacher network,
and the knowledge conveyed by the booster network during the training process.

\begin{table*}[!t]

  \caption{Comparisons of different framework design's result~(test error) on CIFAR-10.}
  \label{tab:ablation}
\centering{{
\begin{threeparttable}
\begin{tabular}{ccccccc}
\toprule
light& booster & base & rocket~(no GB)\tnote{1}& rocket~(no sharing)\tnote{2} & rocket~(no joint training)\tnote{3}& rocket \\
\midrule
WRN-16-1(b) & WRN-40-1 & 8.77 & 8.50  & 8.06 & 8.04 & \textbf{7.87} \\
WRN-16-1(a) & WRN-40-1 & 8.69 & 8.30  & 8.23 & 8.23  & \textbf{7.85}\\
\bottomrule
\end{tabular}
\begin{tablenotes}[para,flushleft]
        \item[1] rocket~(no GB) means rocket launching without gradient block.
        \item[2] rocket~(no sharing) means rocket launching without parameter sharing.
        \item[3] rocket~(no joint training) means booster net trains first, then light net use some layers of booster to initialize, and use hint loss to learn booster net's logits.
      \end{tablenotes}
    \end{threeparttable}}}
\end{table*}
We also investigate our framework with different hint loss formulations.
From Table~\ref{tab:hint}, we see that the adopted hint loss to match the logits achieves the best performance
among the different objectives.
While hint loss to match the probability performs worst, which means gradient vanishing affects the training process.
The experimental results are in accordance with our previous analysis.\par
\subsubsection{Performance of each part of our framework}
\begin{figure}
\centering
\subfigure[bottom rocket net on wide residual net]{
\begin{minipage}[b]{0.45\textwidth}
\includegraphics[width=1\textwidth]{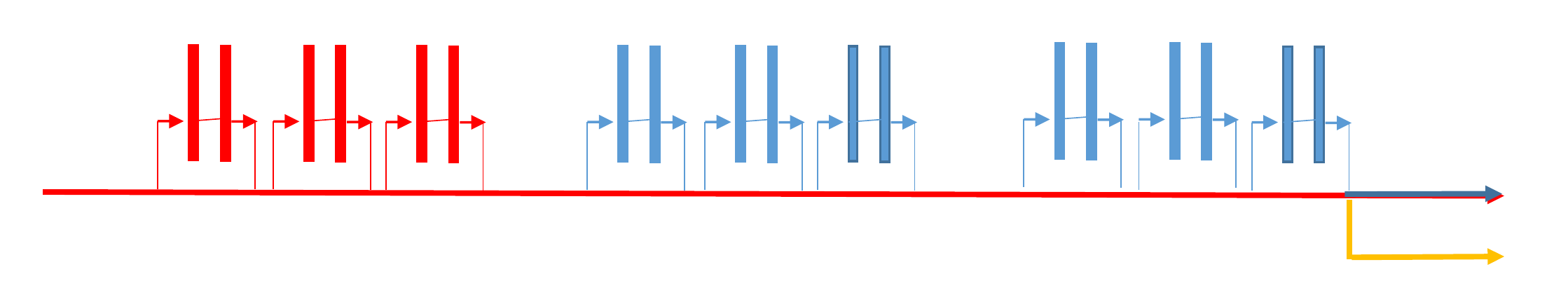}
\end{minipage}
}

\subfigure[interval rocket net on wide residual net]{
\begin{minipage}[b]{0.45\textwidth}
\includegraphics[width=1\textwidth]{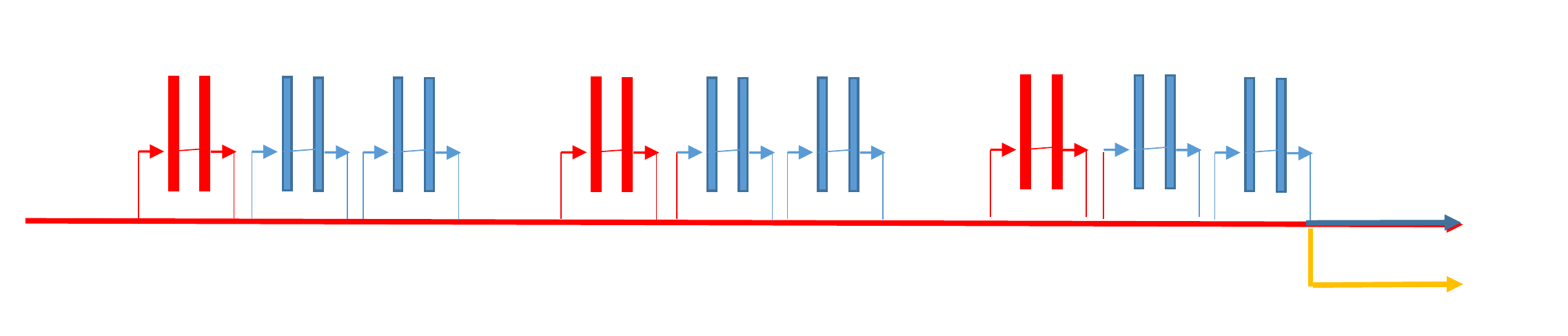}
\end{minipage}
}
\caption{Proposed network structures for rocket net.}
\label{fig:structure}
\end{figure}

Experiments are also carried out on CIFAR-10 to evaluate our framework design~(see Fig.~\ref{fig:structure}).
We observe that simultaneous training, layer sharing and gradient block all contribute to the improvements of our approach. For WRN-16-1(b), gradient block~(GB) gets $0.63\%$ improvement compared with rocket~(no GB); Parameter sharing gets $0.19\%$ improvement compared with rocket~(no sharing). Using part parameter from booster to initialize the light net, using both cross-entropy 
which learns the ground truth
and  $\mathcal{L}_\mathrm{mimic}$ between light net's logits and fixed logits from booster to train light net alone, we get worse results than rocket, which shows the effectiveness of simultaneous training.\par
Besides, our rocket launching with joint training could reduce the whole training time. On CIFAR-10 dataset, the training process of 40-layer booster takes 173 epochs to converge, with 24.6s per epoch on average, and the training of the 16-layer light net takes 165 epochs, with 18.3s for each epoch. The total time is 7275.3s. In contrast, our rocket launching process takes 180 epochs and a total time of 6153.0s to converge, with 34.2s for each epoch. We see that the rocket launching do shorten the time for training the architecture when compared with training the two networks separately.

\begin{table}[!t]
  \caption{Different hint loss functions on CIFAR-10}
  \label{tab:hint}
\centering{\footnotesize
 \centerline {\begin{tabular}{ccccc}
 \toprule
light & booster & $\mathcal{L}_\mathrm{mimic}$ & $\mathcal{L}_\mathrm{MSE}$ & $\mathcal{L}_\mathrm{KD}$ \\
\midrule
WRN-16-1 (b) & WRN-40-1 & 7.87 & 8.32 & \textbf{7.98} \\
WRN-16-1 (a) & WER-40-1 & 7.85 & 8.36 & \textbf{8.26} \\
\bottomrule
    \end{tabular}}}
\end{table}

\subsubsection{Performance with different depths}
In this part, we investigate the learning capability of the light model with different depths and parameter sizes.
Different from previous net structure, in order to make the size of parameter proportional to the layers, we use residual net with fixed width from bottom to top. Light net shares the bottom $n_s$ convolutional layers with booster net. we tune the number of $n_s$ from $10$ to $18$ while the number of booster net's layers is 40(In order to make booster has prominent better learning ability than light net, we set $n_s$ less than half of booster's depth).\par
\begin{figure}[!t]
\begin{minipage}[h]{0mm}
\centering
\includegraphics[width=80mm ]{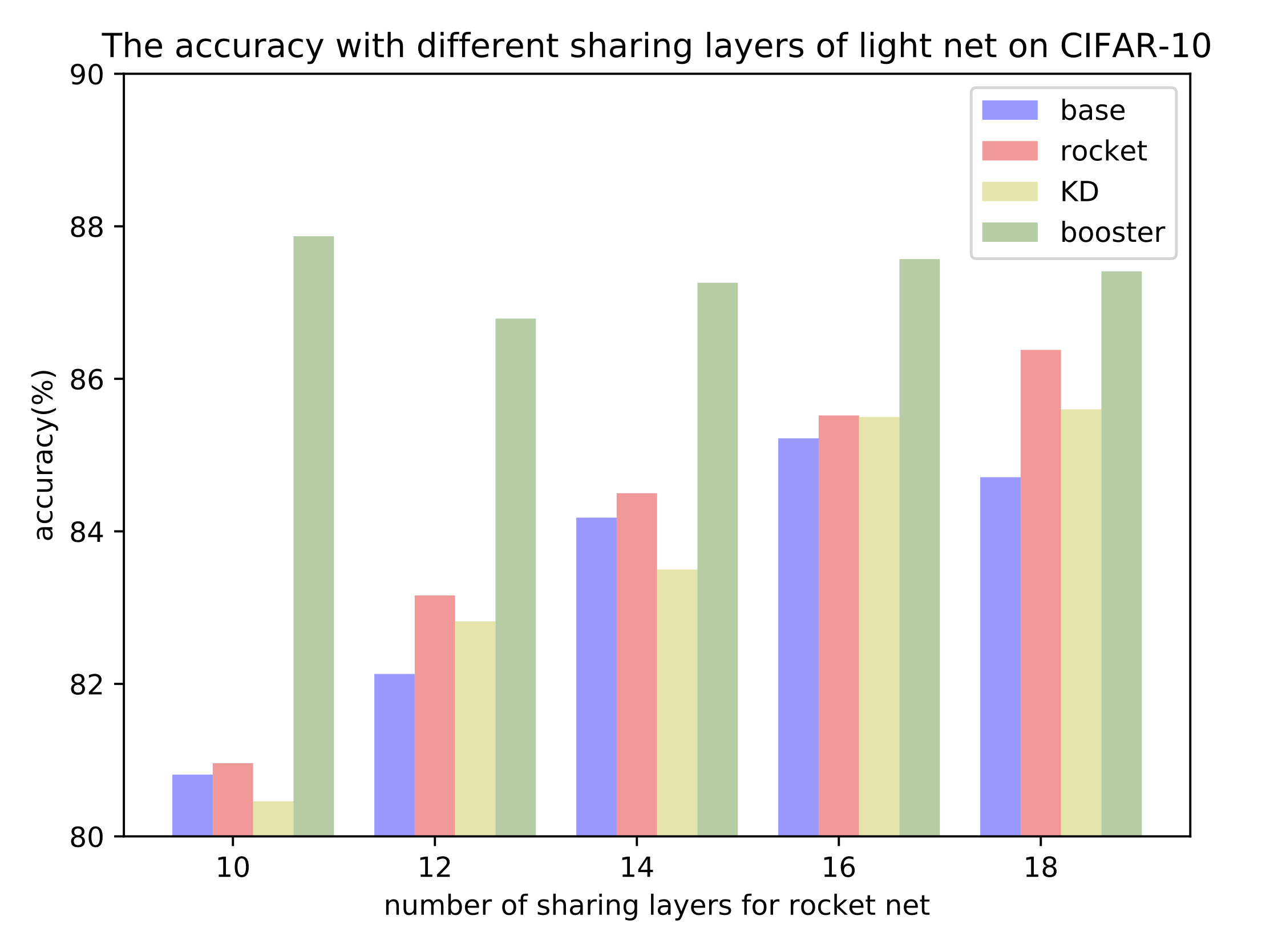}\\
\end{minipage}
\caption{{The accuracy with different sharing layers of light net on CIFAR-10}}
\label{fig:parasize}
\end{figure}

From Fig.~\ref{fig:parasize}, we see that the light model performs better than base and KD stably, which means our light net with different depths all can get extra information with the help of cumbersome booster. It's notable that the gap between base and rocket is not proportional to the depth of light net, this phenomenon may be caused by the balance between light learning ability and extra information from booster net.


\subsubsection{Visualization of rocket launching and attention transfer }
\begin{figure}[!t]
\centering
\subfigure[different group's visualization result on attention transfer]{
\begin{minipage}[htb]{0.45\textwidth}
\includegraphics[width=1\textwidth]{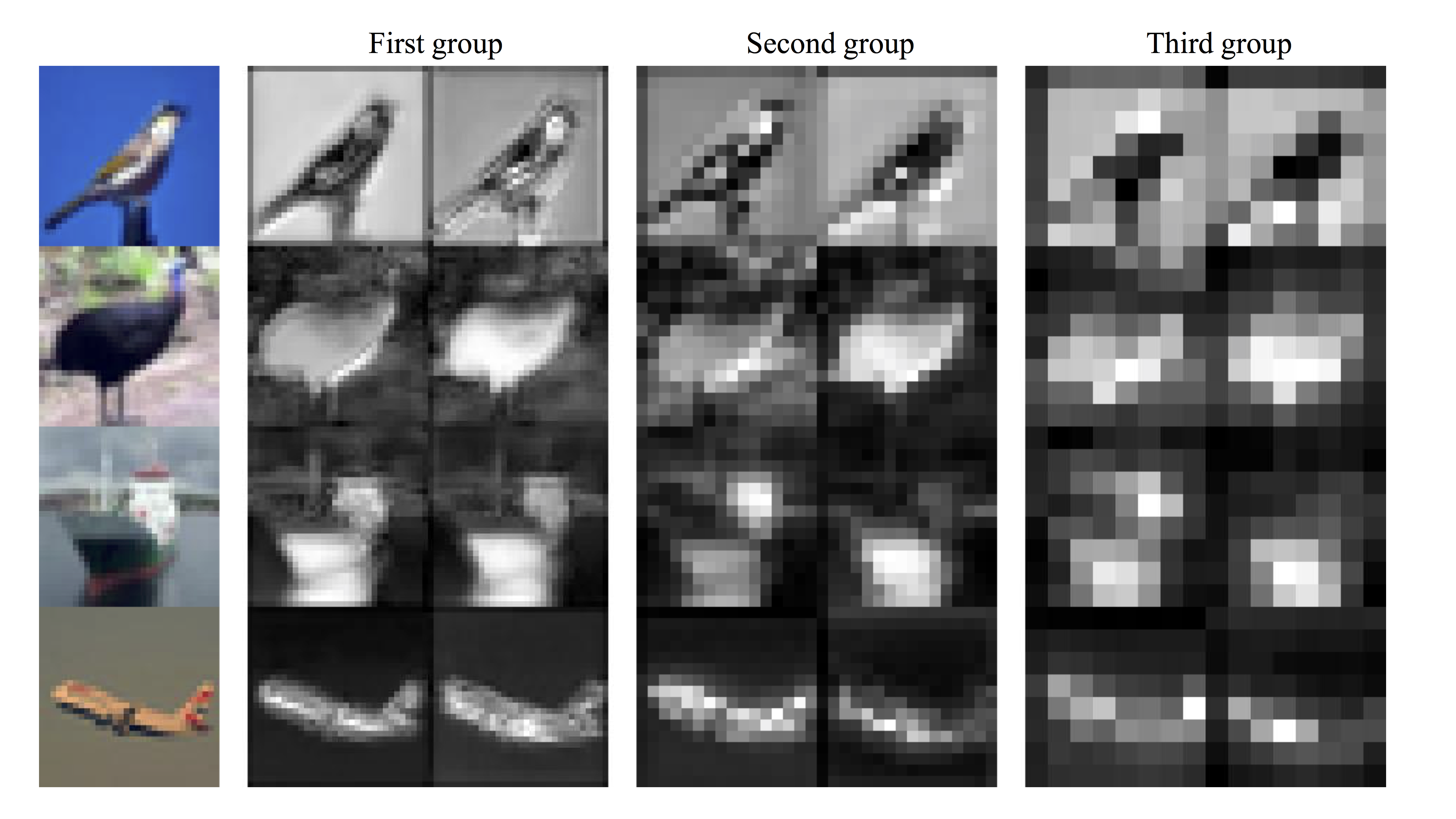}
\end{minipage}}
\subfigure[different group's visualization result on rocket launching]{
\begin{minipage}[htb]{0.45\textwidth}
\includegraphics[width=1\textwidth]{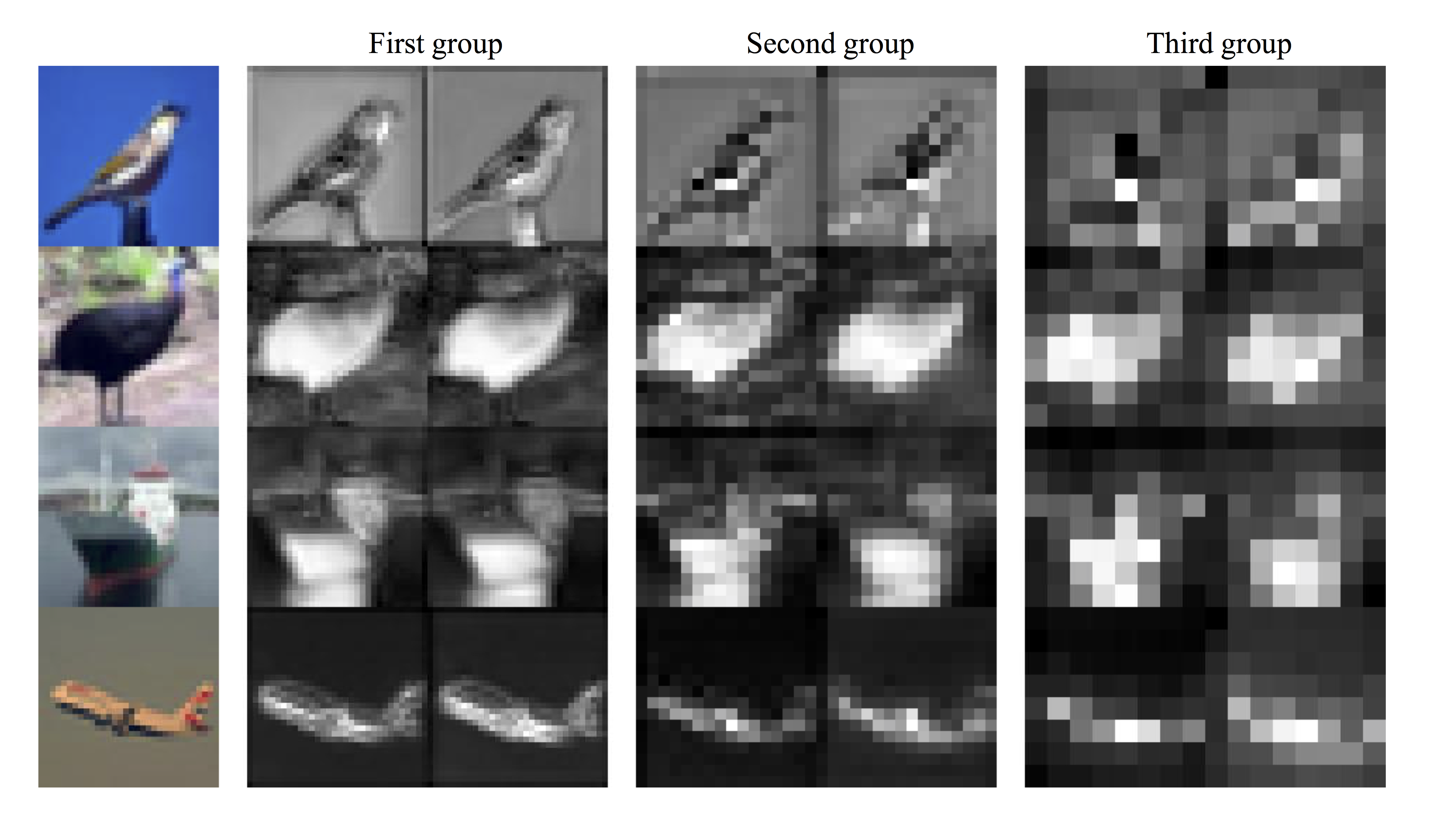}
\end{minipage}
}
\caption{The visualization results on both rocket launching and attention transfer, in each group, the first and second picture in each group stands for booster net and light net respectively }
\label{fig:vis}
\end{figure}

In order to explain our method intuitively, we visualize each group's output of light net and booster net respectively. To be consist with previous part, we use Fig.~\ref{fig:structure}(b) as the basic net. For comparison, we visualize the corresponding results of spatial attention mapping.
As we can see from Fig.~\ref{fig:vis}, for both rocket launching and attention transfer (AT), the feature maps generated from lower groups are similar between light and booster net. It indicates that parameter sharing and attention have similar effect on lower layers. It can also show that these methods can learn the feature representation from booster net in low layer.

\begin{table*}[htbp]
  \caption{Comparisons of classification performance~(test error) on CIFAR-100 and SVHN}
  \label{tab:svhn}
\centering{
\centerline {\begin{tabular}{cccccccc}
\toprule
dataset & light & booster & base & AT & KD & rocket & rocket+KD \\
\midrule
SVHN      & WRN-16-1,~0.2M(b) & WRN-40-1,~0.6M & 3.58 & 2.99 & 2.31 & 2.29 & \textbf{2.20} \\
CIFAR-100 & WRN-16-1,~0.2M(b) & WRN-40-1,~0.6M & 43.7  & 34.1 & 36.4  & 33.3  & \textbf{33.0}  \\
\bottomrule
\end{tabular}}}
\end{table*}

\begin{table*}[htbp]
  \caption{Experiments on real Advertisement Dataset}
  \label{tab:Alibaba}
\centering{\small
 \centerline {\begin{tabular}{ccccc}
\toprule
\ model & \# params in FC layers   & \# multiplications in FC layers & \# inference time of FC Layers & GAUC\\
\midrule
base             & 576 $\times$ 200 $\times$ 80 $\times$ 2  & 131360 & 7.6 ms & 0.632 \\
rocket       & 576 $\times$ 200 $\times$ 80 $\times$ 2  & 131360 & 7.6 ms & 0.635 \\
booster only   & 576 $\times$ 720 $\times$ 360 $\times$ 240 $\times$ 180 $\times$ 90 $\times$ 2  & 837900 & 23.2 ms & 0.637 \\
\bottomrule
\end{tabular}}}
\end{table*}
\subsection{Experiments on SVHN and CIFAR-100}
Aiming to verify the effectiveness of rocket launching further, we apply our method on CIFAR-100 and SVHN respectively. In order to compare with AT (which is based on WRN), we still use WRN as basic net structure, and the sharing method is shown in Fig.~\ref{fig:structure}(b) .

The CIFAR-100 dataset~\cite{krizhevsky2009learning} consists of $32\times32$ color images from $100$ classes.
Like CIFAR-10, these images are still split into $50,000$ training and $10,000$ testing samples. The experiment setting of CIFAR-100 is same as CIFAR-10.\par
The SVHN database~\cite{netzer2011reading} is obtained from house numbers in Google Street View images. It contains $32\times 32$ images with RGB color channels in $10$ class. There are $73,257$ images in the training set, $26,032$ images in testing set and $531,131$ samples in extra set. We follow the same evaluation procedure as Sermanet et al.~\cite{sermanet2012convolutional} to compose our training, validation and test sets. For this dataset, we use validation dataset to choose the final model. In our experiment, we use Adam~\cite{kingma2014adam} with initial learning rate 0.001, while we drop learning rate by $0.2$ at $[20,40,60]$ epochs. Because this dataset is easy to learn, we add dropout after each specific layer of booster net with dropout rate $20\%$ to prevent overfitting. For booster trained alone, same dropout layers are added to keep consistent.

Error rate on above two dataset is shown in Table~\ref{tab:svhn}. We observe that our approach gets $1.29\%$ improvement on SVHN and $10.4\%$ improvement on CIFAR-100 compared with base model. What's more, rocket launching outperforms other teacher-student methods on all settings.

\subsection{Experiments on real Advertisement Dataset}
In order to verity the effectiveness of rocket launching further, we test our method on huge real industry dataset. The dataset\footnote{ https://tianchi.aliyun.com/datalab/dataSet.htm?spm=5176100\par073.888.26.70c5adaeMeJQpW\&id=19} comes from productive display advertising system in Alibaba, we use rocket launching to predict whether user clicks given product. The size of training set is $4$ billion, the test set is $0.285$ billion.\par
The network that we use is shown in DIN~\cite{zhou2017deep}.
In the online system, most calculations focus on the fully connected layers after the embedding layers. So we try to use a booster net with more complex fully connected layers to guide our light net. The light net shares embedding layers with booster net. The booster net has seven wide hidden layers using complicated operation like batch normalization\cite{BatchNorm}, light net's specific layers with less hidden units and has only fully connected layers. The light net in the huge real data gets 0.3\% improvement on GAUC (the generalization of AUC)~\cite{zhou2017deep} with the same latency as the base model. The booster net gets the best performance on the offline metric,
but it needs $23.2$ ms for one requirement to infer hundreds candidate advertisements,
which is unacceptable for online system. Our approach can get improvement on a model with same structure and parameter quantity. And this experiment proves that one can use our approach to break the boundary brought by the latency limitation to some degree.

\section{Conclusion}
We propose a general framework named rocket launching to get a efficient well-performing light model with the help of a cumbersome booster net.
In order to get as much as information from the booster model, we make the booster and the light net train on the same task together with the hint loss objective,
which pushes the booster model to supervise the whole training process of the light one.
Besides, the light model shares parameter with the booster to make the light net get low-level representation directly from the booster.
We also analyze different hint loss functions cludethat can convey knowledge from the booster to the light model.
Moreover, we develop the gradient block scheme to prevent the booster net from deterioration.
For future work, we would like to explore training networks with not only smaller depths but also fewer neurons
in each layer to further improve the inference efficiency.


\bibliography{reference}

\begin{thebibliography}{}

\bibitem[\protect\citeauthoryear{Andrew \bgroup et al\mbox.\egroup
  }{2013}]{Andrew13dcca}
Andrew, G.; Arora, R.; Bilmes, J.; and Livescu, K.
\newblock 2013.
\newblock Deep canonical correlation analysis.
\newblock In {\em Proceedings of the 30th International Conference on Machine
  Learning},  1247--1255.

\bibitem[\protect\citeauthoryear{Ba and Caruana}{2014}]{ba2014deep}
Ba, J., and Caruana, R.
\newblock 2014.
\newblock Do deep nets really need to be deep?
\newblock In {\em Advances in neural information processing systems 27},
  2654--2662.
\newblock Cambridge, MA.: MIT Press.

\bibitem[\protect\citeauthoryear{Bahdanau, Cho, and
  Bengio}{2014}]{bahdanau2014neural}
Bahdanau, D.; Cho, K.; and Bengio, Y.
\newblock 2014.
\newblock Neural machine translation by jointly learning to align and
  translate.
\newblock {\em arXiv preprint arXiv:1409.0473}.

\bibitem[\protect\citeauthoryear{Bromley \bgroup et al\mbox.\egroup
  }{1994}]{Bromley1993siamese}
Bromley, J.; Guyon, I.; LeCun, Y.; S{\"a}ckinger, E.; and Shah, R.
\newblock 1994.
\newblock Signature verification using a siamese time delay neural network.
\newblock In {\em Proceedings of the 12th Advances in Neural Information
  Processing Systems},  737--744.

\bibitem[\protect\citeauthoryear{Bucilu\v{a}, Caruana, and
  Niculescu-Mizil}{2006}]{Bucilua06compression}
Bucilu\v{a}, C.; Caruana, R.; and Niculescu-Mizil, A.
\newblock 2006.
\newblock Model compression.
\newblock In {\em Proceedings of the 12th ACM SIGKDD international conference
  on Knowledge discovery and data mining},  535--541.

\bibitem[\protect\citeauthoryear{Cheng \bgroup et al\mbox.\egroup
  }{2016}]{cheng2016wide}
Cheng, H.-T.; Koc, L.; Harmsen, J.; Shaked, T.; Chandra, T.; Aradhye, H.;
  Anderson, G.; Corrado, G.; Chai, W.; Ispir, M.; et~al.
\newblock 2016.
\newblock Wide \& deep learning for recommender systems.
\newblock In {\em Proceedings of the 1st Workshop on Deep Learning for
  Recommender Systems},  7--10.
\newblock Boston, USA: ACM.

\bibitem[\protect\citeauthoryear{Denton \bgroup et al\mbox.\egroup
  }{2014}]{denton2014exploiting}
Denton, E.~L.; Zaremba, W.; Bruna, J.; LeCun, Y.; and Fergus, R.
\newblock 2014.
\newblock Exploiting linear structure within convolutional networks for
  efficient evaluation.
\newblock In {\em Advances in Neural Information Processing Systems 27},
  1269--1277.
\newblock Cambridge, MA.: MIT Press.

\bibitem[\protect\citeauthoryear{He \bgroup et al\mbox.\egroup
  }{2016}]{he2016deep}
He, K.; Zhang, X.; Ren, S.; and Sun, J.
\newblock 2016.
\newblock Deep residual learning for image recognition.
\newblock In {\em Proceedings of the IEEE conference on computer vision and
  pattern recognition},  770--778.

\bibitem[\protect\citeauthoryear{He \bgroup et al\mbox.\egroup
  }{2017}]{He2017maskrcnn}
He, K.; Gkioxari, G.; Dollar, P.; and Girshick, R.
\newblock 2017.
\newblock Mask r-cnn.
\newblock {\em arXiv preprint arXiv:1703.06870}.

\bibitem[\protect\citeauthoryear{Hinton, Vinyals, and
  Dean}{2015}]{hinton2015distill}
Hinton, G.; Vinyals, O.; and Dean, J.
\newblock 2015.
\newblock Distilling the knowledge in a neural network.
\newblock {\em arXiv preprint arXiv:1503.02531}.

\bibitem[\protect\citeauthoryear{Howard \bgroup et al\mbox.\egroup
  }{2017}]{howard2017mobilenets}
Howard, A.~G.; Zhu, M.; Chen, B.; Kalenichenko, D.; Wang, W.; Weyand, T.;
  Andreetto, M.; and Adam, H.
\newblock 2017.
\newblock Mobilenets: Efficient convolutional neural networks for mobile vision
  applications.
\newblock {\em arXiv preprint arXiv:1704.04861}.

\bibitem[\protect\citeauthoryear{Huang \bgroup et al\mbox.\egroup
  }{2016}]{huang2016densely}
Huang, G.; Liu, Z.; Weinberger, K.~Q.; and van~der Maaten, L.
\newblock 2016.
\newblock Densely connected convolutional networks.
\newblock {\em arXiv preprint arXiv:1608.06993}.

\bibitem[\protect\citeauthoryear{Ioffe and Szegedy}{2015}]{BatchNorm}
Ioffe, S., and Szegedy, C.
\newblock 2015.
\newblock Batch normalization: Accelerating deep network training by reducing
  internal covariate shift.
\newblock {\em CoRR} abs/1502.03167.

\bibitem[\protect\citeauthoryear{Kingma and Ba}{2014}]{kingma2014adam}
Kingma, D., and Ba, J.
\newblock 2014.
\newblock Adam: A method for stochastic optimization.
\newblock {\em arXiv preprint arXiv:1412.6980}.

\bibitem[\protect\citeauthoryear{Krizhevsky and
  Hinton}{2009}]{krizhevsky2009learning}
Krizhevsky, A., and Hinton, G.
\newblock 2009.
\newblock Learning multiple layers of features from tiny images.

\bibitem[\protect\citeauthoryear{Krizhevsky, Sutskever, and
  Hinton}{2012}]{krizhevsky2012imagenet}
Krizhevsky, A.; Sutskever, I.; and Hinton, G.~E.
\newblock 2012.
\newblock Imagenet classification with deep convolutional neural networks.
\newblock In {\em Advances in neural information processing systems 25},
  1097--1105.

\bibitem[\protect\citeauthoryear{Laine and Aila}{2016}]{laine2016temporal}
Laine, S., and Aila, T.
\newblock 2016.
\newblock Temporal ensembling for semi-supervised learning.
\newblock {\em arXiv preprint arXiv:1610.02242}.

\bibitem[\protect\citeauthoryear{Luo, Wu, and Lin}{2017}]{luo2017thinet}
Luo, J.-H.; Wu, J.; and Lin, W.
\newblock 2017.
\newblock Thinet: A filter level pruning method for deep neural network
  compression.
\newblock {\em arXiv preprint arXiv:1707.06342}.

\bibitem[\protect\citeauthoryear{Netzer \bgroup et al\mbox.\egroup
  }{2011}]{netzer2011reading}
Netzer, Y.; Wang, T.; Coates, A.; Bissacco, A.; Wu, B.; and Ng, A.~Y.
\newblock 2011.
\newblock Reading digits in natural images with unsupervised feature learning.
\newblock In {\em NIPS workshop on deep learning and unsupervised feature
  learning}, volume 2011, ~5.

\bibitem[\protect\citeauthoryear{Romero \bgroup et al\mbox.\egroup
  }{2014}]{Romero2014fitnet}
Romero, A.; Ballas, N.; Kahou, S.~E.; and Chassang, A.
\newblock 2014.
\newblock Fitnets: Hints for thin deep nets.
\newblock {\em arXiv preprint arXiv:1412.655d0}.

\bibitem[\protect\citeauthoryear{Sermanet, Chintala, and
  LeCun}{2012}]{sermanet2012convolutional}
Sermanet, P.; Chintala, S.; and LeCun, Y.
\newblock 2012.
\newblock Convolutional neural networks applied to house numbers digit
  classification.
\newblock In {\em Pattern Recognition (ICPR), 2012 21st International
  Conference on},  3288--3291.
\newblock IEEE.

\bibitem[\protect\citeauthoryear{Simonyan and
  Zisserman}{2015}]{Simonyan2015verydeep}
Simonyan, K., and Zisserman, A.
\newblock 2015.
\newblock Very deep convolutional networks for large-scale image recognition.
\newblock In {\em Proceedings of the 3th International Conference on Learning
  Representations},  621--630.

\bibitem[\protect\citeauthoryear{Szegedy \bgroup et al\mbox.\egroup
  }{2015}]{szegedy2015going}
Szegedy, C.; Liu, W.; Jia, Y.; Sermanet, P.; Reed, S.; Anguelov, D.; Erhan, D.;
  Vanhoucke, V.; and Rabinovich, A.
\newblock 2015.
\newblock Going deeper with convolutions.
\newblock In {\em Proceedings of the IEEE conference on computer vision and
  pattern recognition},  1--9.

\bibitem[\protect\citeauthoryear{Zagoruyko and
  Komodakis}{2016}]{Zagoruyko2016attenion}
Zagoruyko, Z., and Komodakis, K.
\newblock 2016.
\newblock Paying more attention to attention: Improving the performance of
  convolutional neural networks via attention transfer.
\newblock {\em arXiv preprint arXiv:1612.03928}.

\bibitem[\protect\citeauthoryear{Zagoruyko and Nikos}{2016}]{Zagoruyko2016wide}
Zagoruyko, S., and Nikos, K.
\newblock 2016.
\newblock Wide residual networks.
\newblock {\em arXiv preprint arXiv:1605.07146}.

\bibitem[\protect\citeauthoryear{Zhang \bgroup et al\mbox.\egroup
  }{2017}]{zhang2017shufflenet}
Zhang, X.; Zhou, X.; Lin, M.; and Sun, J.
\newblock 2017.
\newblock Shufflenet: An extremely efficient convolutional neural network for
  mobile devices.
\newblock {\em arXiv preprint arXiv:1707.01083}.

\bibitem[\protect\citeauthoryear{Zhou \bgroup et al\mbox.\egroup
  }{2017}]{zhou2017deep}
Zhou, G.; Song, C.; Zhu, X.; Ma, X.; Yan, Y.; Dai, X.; Zhu, H.; Jin, J.; Li,
  H.; and Gai, K.
\newblock 2017.
\newblock Deep interest network for click-through rate prediction.
\newblock {\em arXiv preprint arXiv:1706.06978}.

\bibitem[\protect\citeauthoryear{Zhou}{2016}]{zhou2016learnware}
Zhou, Z.-H.
\newblock 2016.
\newblock Learnware: on the future of machine learning.
\newblock {\em Frontiers of Computer Science} 10(4):589--590.

\end{thebibliography}
\bibliographystyle{aaai}
\end{document}